%% file: template.tex
\documentclass{article}

\usepackage{arxiv}

\usepackage[utf8]{inputenc} 
\usepackage[T1]{fontenc}    
\usepackage{hyperref}       
\usepackage{url}            
\usepackage{booktabs}       
\usepackage{amsfonts}       
\usepackage{nicefrac}       
\usepackage{microtype}      
\usepackage{graphicx}
\usepackage{natbib}
\usepackage{doi}

\usepackage{url}
\usepackage{xcolor}
\usepackage{subfig}
\usepackage{graphicx}
\usepackage{amsmath}
\usepackage{multirow}
\usepackage{latexsym}

\title{Leveraging counterfactual concepts for debugging and improving CNN model performance \thanks{This manuscript is currently under consideration for publication in \textit{Pattern Recognition Letters}.}
}

\date{} 					

\author{ Syed Ali Tariq \\
	Department of Computer Science\\
	COMSATS University Islamabad\\
	Islamabad, Pakistan \\
	\texttt{s.alitariq1@gmail.com} \\
	\And
	Tehseen Zia \\
	Department of Computer Science\\
	COMSATS University Islamabad\\
	Islamabad, Pakistan \\
	\texttt{tehseen.zia@comsats.edu.pk} \\
}

\begin{document}
\maketitle

\begin{abstract}
	Counterfactual explanation methods have recently received significant attention for explaining CNN-based image classifiers due to their ability to provide easily understandable explanations that align more closely with human reasoning. However, limited attention has been given to utilizing explainability methods to improve model performance. In this paper, we propose to leverage counterfactual concepts aiming to enhance the performance of CNN models in image classification tasks. Our proposed approach utilizes counterfactual reasoning to identify crucial filters used in the decision-making process. Following this, we perform model retraining through the design of a novel methodology and loss functions that encourage the activation of class-relevant important filters and discourage the activation of irrelevant filters for each class. This process effectively minimizes the deviation of activation patterns of local predictions and the global activation patterns of their respective inferred classes. By incorporating counterfactual explanations, we validate unseen model predictions and identify misclassifications. The proposed methodology provides insights into potential weaknesses and biases in the model's learning process, enabling targeted improvements and enhanced performance. Experimental results on publicly available datasets have demonstrated an improvement of 1-2\%, validating the effectiveness of the approach.
\end{abstract}

\keywords{Explainable AI \and misclassification detection \and counterfactual explanation \and model debugging}

\input{intro}

\input{Literature}
\input{methodology}

\input{results}

\input{conclusion}

\bibliographystyle{plainnat}
\bibliography{bibliography}






\end{document}

%% file: intro.tex
\section{Introduction}

Deep learning (DL) models have achieved remarkable progress in computer vision, particularly in image classification tasks using convolutional neural network (CNN) \cite{krizhevsky2012imagenet, he2016deep, tan2019efficientnet}. However, their lack of transparency and interpretability remains a challenge \cite{rudin2019stop, arrieta2020explainable}. To tackle this issue, various explainable artificial intelligence (XAI) techniques have been introduced that aim to explain the decisions of DL models. These techniques include saliency maps \cite{wang2019learning}, activation visualization \cite{selvaraju2017grad}, concept attribution \cite{wu2020towards}, and counterfactual explanations (CEs) \cite{wang2020scout}, among others. CEs have emerged as a promising approach for explaining CNN-based image classifiers due to their ability to provide easy-to-understand explanations that are aligned more closely with the way humans reason \cite{byrne2019counterfactuals}.

Despite these advancements, limited attention has been given to leveraging explanations and model interpretability to enhance model performance and detect potential data biases. In various domains such as medical diagnosis, autonomous driving, and facial recognition, the need for highly accurate and explainable DL models is crucial, which can impact their trustworthiness and adoption in real-world settings \cite{rudin2019stop}. It is essential to explore novel methods that provide both accuracy and interpretability in order to improve model performance and trustworthiness.

Several recent works have attempted to diagnose and improve the performance of DL models \cite{feng2022model, leclerc20223db, abid2022meaningfully, alsallakh2021debugging}. These methods aim to identify and rectify issues within DL models by analyzing their internal components and examining the sources of errors. In a recent study \cite{TARIQ2022109901}, the authors proposed a causal reasoning process to identify counterfactual filters within CNNs, revealing the intrinsic decision-making process of CNN models. While these advancements have contributed to the field of XAI, the potential of leveraging explanations and model interpretability to enhance model performance and detect biases has been relatively overlooked.

In this paper, we aim to bridge this gap by proposing an approach that leverages counterfactual concepts to enhance CNN model performance. We utilize a previously developed counterfactual filter identification model \cite{TARIQ2022109901} to identify key filters used in the decision-making process of images. We extend this approach to identify class-specific important filters that contribute the most towards the prediction of images to respective classes. Using the identified class-specific filters, we find out the sources of errors and biases in the model's decision for unseen images by analyzing the filter activation patterns and how much they align or deviate from the activation patterns of inferred classes. This technique allows us to perform misclassification analysis to detect instances where the model makes incorrect predictions and understand why the model is making these mistakes. This information can then be used to improve the model's accuracy by addressing the specific issues that are causing the misclassifications. By analyzing misclassifications and identifying patterns or biases in the model's decision-making process, we develop a strategy to improve the model's overall performance. We achieve this by proposing a novel training technique and designing loss functions that encourage the activation of class-relevant important filters and discourage the activation of irrelevant filters for each class, resulting in less deviation of activation patterns of filters in local predictions in comparison to key filters identified for each class.

Our proposed method demonstrates its effectiveness by improving the false-positive rate of a trained model by 1-2\%. This improvement is especially significant in critical domains such as healthcare and finance, where even a single misclassification can have severe consequences. The contributions of this work are twofold: (1) conducting thorough misclassification analysis to assess the likelihood of correct predictions and (2) retraining the model based on misclassifications by aligning the decision-making process with class-specific key filters.

Our proposed method improves the false-positive rate of a trained model by 1-2\%, demonstrating the effectiveness of the approach. This is especially important in critical domains such as healthcare and finance, where even a single misclassification can have severe consequences. The key contributions of this work are twofold: (1) developing a methodology to conduct misclassification analysis to assess the likelihood of correct predictions, and (2) developing a training strategy to improve pre-trained models by aligning the decision-making process of the model to class-specific key filters.

The rest of the paper is organized as follows. Section \ref{sec:two} provides an overview of related work in the field. In Section \ref{sec:three}, we present our proposed methodology. The experimental results are discussed in Section \ref{sec:four}, and finally, Section \ref{sec:five} concludes the paper.

%% file: Literature.tex
\section{Related Work}
\label{sec:two}

In the literature, there exist several XAI and model debugging techniques for analyzing and improving Deep Learning (DL) models \cite{feng2022model, leclerc20223db, abid2022meaningfully, alsallakh2021debugging, TARIQ2022109901}. A brief overview of these works is as follows.
In \cite{feng2022model}, the authors developed a method called "Model Doctor" that automatically diagnoses and treats CNN models. They accomplish this by finding correlations between predicted classes and the convolution kernels active in the decision-making process. The authors conclude that only a sparse set of filters from deeper layers are mostly responsible for making accurate predictions. They propose treating the classifier by aggregating gradients to find correlations between classes and filters and applying constraints on filters to minimize incorrect activations. This leads to an improved performance of 1-5\%, as reported by the authors.

In \cite{leclerc20223db}, a debugging framework for computer vision models is proposed. It allows users to select and configure various transformations on the inputs to the model. The framework then performs analysis on the input space to identify failure cases and conducts global as well as per-object analysis.

In \cite{abid2022meaningfully}, the authors propose a counterfactual concepts-based debugging method for classifiers. This method learns important concepts from limited training data and uses them to validate misclassified samples. It assigns different scores to different concepts, which would lead to the model correcting its errors.

In \cite{alsallakh2021debugging}, the authors explore the internal workings of convolutional neural networks (CNNs) to debug models. They analyze convolution operations and find that they can result in artifacts and noise appearing on feature maps, as well as weights of the convolutional filters. The authors present a visual debugging method to visualize these sources of errors and how they affect the desired output of the model. They also discuss different strategies to mitigate such errors.

In \cite{TARIQ2022109901}, the authors explore the internal workings of CNN classifiers and propose a counterfactual and contrastive reasoning method based on the convolutional filters used in the decision-making process. They extract important filters that contribute the most to a particular prediction and also extract filters that, if altered, would lead the model to change its prediction to another target class. Using these filters, the authors visually demonstrate the important concepts being learned by the classifier and how they affect the prediction output.

These methods aim to identify and rectify issues within DL models by analyzing their internal components and examining the sources of errors. However, these existing model debugging methods have certain limitations. They often rely on complex procedures that may be computationally expensive, require large amounts of annotated data, or lack generalizability across different models and datasets. These limitations highlight the need for further research and development of effective debugging techniques.

%% file: methodology.tex
\section{Proposed methodology}
\label{sec:three}

The objective of the proposed work is to identify biases in the model decisions that lead to misclassifications and develop a model debugging method to improve model performance on misclassifications. To accomplish these goals, we build upon the existing counterfactual explanation (CFE) model that identifies counterfactual filters to explain model decisions \cite{TARIQ2022109901}. The CFE model works by predicting a set of minimum correct (MC) and minimum incorrect (MI) filters necessary to (i) maintain the prediction of the image to the original inferred class by the classifier and (ii) alter the classifier's decision to a chosen target class. 


\begin{figure*}[hbt!]
\begin{center}
    \includegraphics[width=1.0\linewidth]{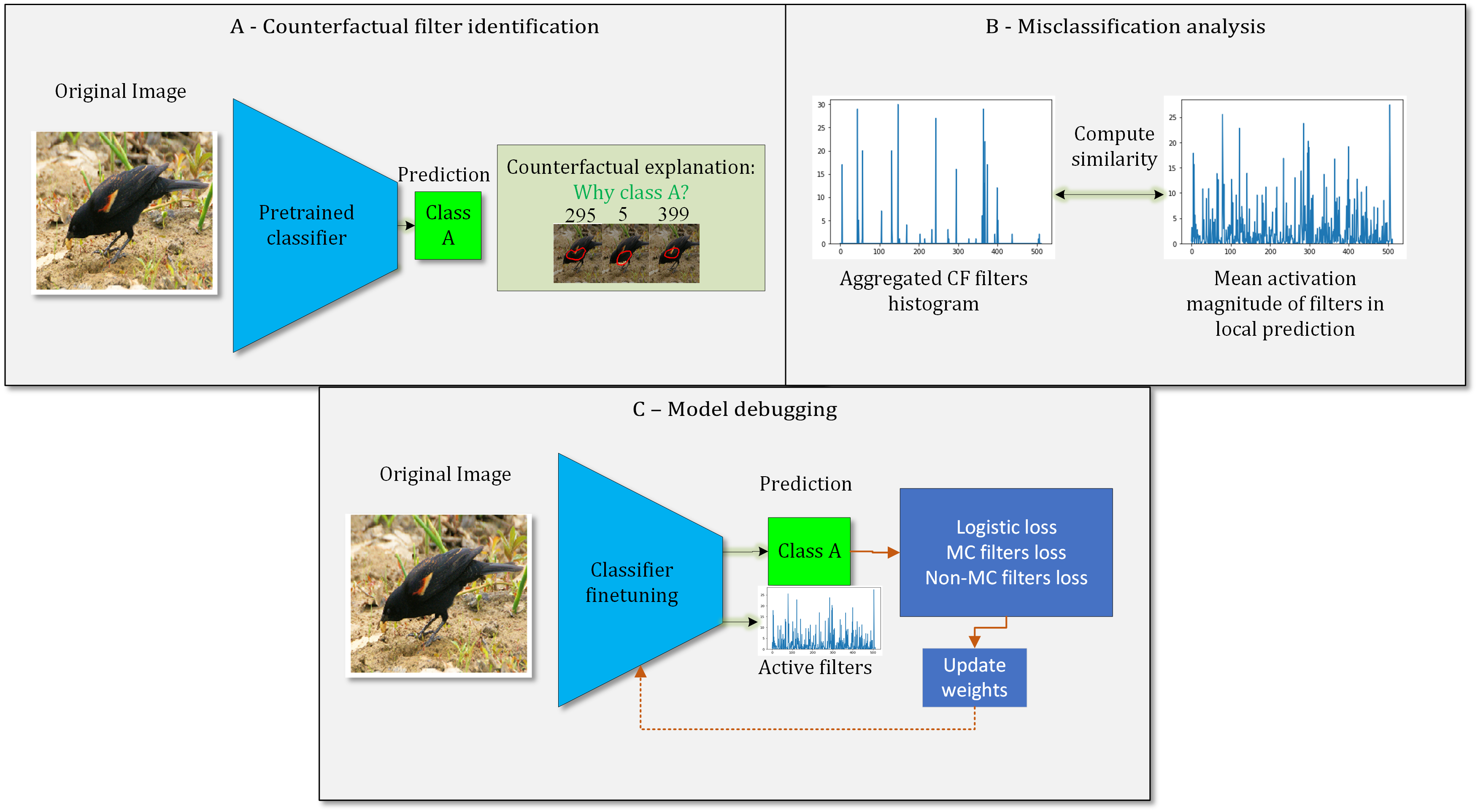}
\end{center}
   \caption{Block diagram of the proposed misclassification detection and model debugging method.}
\label{fig:1}
\end{figure*}

In the proposed work, we use the MC filters to identify the globally most important filters for training images, enabling us to perform misclassification analysis and model debugging on unseen test images. Fig. \ref{fig:1} presents a block diagram of our proposed model, illustrating the various phases involved. In the counterfactual filter identification phase, we extract the MC filters to provide counterfactual explanations for the classifier's decisions. Subsequently, in the misclassification analysis phase, we accumulate the class-specific filters identified for each training image, along with their respective inferred class if it is the true class. This global class-specific filter set represents each class's most important filters.
To detect misclassifications in new test images, we measure the level of similarity or agreement between the activated MC filters for the test images and their inferred class, comparing them with the corresponding global MC filters. By assessing the degree of agreement, we can determine whether the inferred class is likely to be a misclassification or not. A higher level of agreement indicates a more reliable prediction for the inferred class, as it confirms that the features extracted from the current test image align with what the model has learned from the training images of that class. This approach allows us to detect misclassifications by identifying abnormal filter activation patterns that deviate from the observed activation patterns in the training images, providing insights into biases within the model's decision-making process. Finally, in the last phase, we perform model treatment to debug the model's performance. This involves retraining the model using novel loss functions designed to encourage the activation of the globally important filters and discourage the activation of irrelevant filters for each class.



The equation for accumulating the MC filters for each training image is as follows. For a given input image $x_i \in X$, we compute the predicted class $\hat{c}$ using the classifier $C$. 
\begin{equation}
    \hat{c_i} = \arg\max (C(x_i))
\end{equation}
If the $\hat{c}$ is the same as the true class, and the classifier is highly confident in its decision, i.e., prediction confidence is above a given threshold $\tau$, then we compute and accumulate the local MC filters, represented by $MC_{x_i}$:

\begin{equation}
\label{eq:5}
    \text{If } (\hat{c_i} == c_i) \land (P(\hat{c_i}|x_i) > \tau), \text{ then } MC_c \leftarrow MC_c \cup MC_{x_i}
\end{equation}
where $\hat{c_i}$ represents the predicted class label, $c_i$ represents the ground truth class label, $P(\hat{c_i}|x_i)$ represents the predicted probability of the predicted class given the input $x_i$, $\tau$ represents the specified threshold, ${MC_c}$ represents the global MC filters set for class $c_i$, and ${MC_{x_i}}$ represents the local MC filters for a specific image computed using CFE model as follows:
\begin{equation}
\label{eq:2}
MC_{x_i} = \text{CFE}(x_i, C, \hat{c_i}), 
\end{equation}
The condition Eq. \ref{eq:5} checks whether both the predicted class $\hat{c_i}$ and the predicted probability $P(\hat{c_i}|x_i)$ satisfy the specified conditions. If they do, the local MC filters ($MC_{x_i}$) are accumulated into the global MC filter set (${MC_c}$).
The CFE model works on the last convolution layer of the classifier since these filters are most impactful and meaningful that learn more abstract, high-level features, concepts, and even whole objects as well \cite{bau2020understanding, zhou2014object, bau2017network}. 


Once the globally important class-relevant filters, ${MC_c}$, have been identified for each class, we proceed to treat and debug the model by encouraging the activation of class-relevant global filters while discouraging the activation of filters important for other classes. To accomplish this, we formulate the following loss functions that guide the retraining process.

During the treating phase, we utilize all available training images to retrain the classifier $C$ with two additional loss functions in addition to the cross-entropy loss. To encourage activation of MC filters, we maximize the agreement between global MC filters $MC_c$ and local activated filters $f_{x_i}$ using the following loss function:
\begin{equation}
    L_{MC}(MC_c, f_{x_i}) = -\sum_{k=1}^{n}||MC_c(k) * {f_{x_i}(k)}||,
\end{equation}
where $f_{x_i}$ is a binary filter map representing activated filters in the prediction of image $x_i$, computed as:
\begin{equation}
f_{x_i} = \text{ReLU}_\text{t}(S(g_i)),
\end{equation}
where $g_i$ represents the feature maps after the global average pooling layer of classifier $C(x_i)$, $S$ represents the sigmoid activation function, and ReLU$_\text{t}$ is a thresholded-ReLU layer with threshold $t$ set to $t = 0.5$ that outputs the approximated local binary filter activation map, indicated whether a filter is active or inactive. 

To discourage activation of all other non-MC filters, we minimize the agreement between non-MC global filters $1-MC_c$ and local activated filters $g_{x_i}$:
\begin{equation}
    L_{nonMC}(MC_c, f_{x_i}) = \sum_{k=1}^{n}||(1-MC_c(k)) * {f_{x_i}(k)}||.
\end{equation}
When combined with the cross-entropy loss, represented by $L_{CE}$,
\begin{equation}
    L_{CE}(\hat{c}_{i}, c_i) = -\frac{1}{m}\sum_{i=1}^{m} [c_i\log \hat{c}_{i} + (1-c_i)\log (1-\hat{c}_{i})],
 \end{equation}
where $\hat{c}_{i}$ is the predicted class, $c_i$ is the true class, and $m$ is the total number of training image,
the final loss equation for model debugging, represented by $L_d$, becomes:
\begin{equation}
\label{eq:l_d}
    L_d = L_{CE} - \lambda_1 L_{MC} +  \lambda_2 L_{nonMC},
 \end{equation}
where $\lambda_1$ and $\lambda_2$ are weights assigned to the loss maximizing the agreement with MC filters and minimizing agreement with non-MC filters, respectively. When all three losses are reduced simultaneously, this approach pushes the model to focus only on crucial features for classifying each image to its respective class while minimizing the impact of misleading or irrelevant ones. Through this model debugging technique, we aim to improve the model's overall performance by effectively addressing errors, biases, and misclassifications.

%% file: results.tex
\section{Results}
\label{sec:four}

This section presents the results and discussion of the proposed model debugging method, organized as follows. The experimental setup is described in section \ref{sec:four-1}. In section \ref{sec:four-2}, we identify the class-relevant filters using the CFE model and perform automatic misclassification identification. Model treatment is presented in section \ref{sec:four-4}, where the proposed model debugging method is applied to re-train and improve the model. 
Detailed analysis and interpretation of the results are presented in section \ref{sec:four-5}, while section \ref{sec:four-6} presents a closing discussion on the results section.

\subsection{Experimental setup}
\label{sec:four-1}
For the evaluation of the proposed model debugging method, we used VGG-16 \cite{simonyan2014very} model trained on the Caltech-UCSD Birds (CUB) 2011 \cite{WahCUB_200_2011} dataset. The VGG-16 model was originally trained by transfer learning using imageNet \cite{ILSVRC15} pre-trained weights followed by fine-tuning and achieved training and testing accuracy of 99.0\% and 69.5\%, respectively. 

To debug the VGG-16 model, we utilize the counterfactual explanation model (CFE) \cite{TARIQ2022109901}, which provides contrastive and counterfactual explanations of the decisions made by a pre-trained model. The CFE model predicts the minimum correct (MC) and minimum incorrect (MI) filters for each decision by a pre-trained model with respect to the original inferred class (source class) and some target alter class. In the proposed work, we build upon the CFE model work to detect weak and faulty filters in the pre-trained VGG-16 model that lead to dataset bias and misclassifications and rectify these issues to improve model accuracy. The CFE model creation and training details are provided in \cite{TARIQ2022109901}, and we use the best performing CFE model with hyperparameters defined in \cite{TARIQ2022109901}, i.e., we use the CFE model with logits loss enabled and with sparsity loss value of $\lambda = 2$.


\subsection{Misclassification analysis}

\label{sec:four-2}


The identification of class-relevant important filters is performed using the CFE model that is used to predict the MC filters from the VGG-16 model for each training image of a class. To ensure only high-quality and class-relevant filters are detected, we skip misclassifications and low confidence predictions having probability score of $p<90\%$. The activation magnitudes of the predicted MC filters are accumulated and normalized for each training image of a class. These global MC filters represent the most crucial features or concepts relevant to their respective classes. It is shown in \cite{TARIQ2022109901} that disabling these MC filters significantly reduces the class recall of their classes without notably affecting the overall model accuracy.



Following this, we perform  misclassification analysis on individual test images passed to the CFE model to compute the MC filters for the respective inferred classes. The MC filters for each prediction to the inferred class are compared with the globally important class-specific filters to compute an agreement between the predicted MC filters and class-specific filters. The higher the agreement between the local and global MC filters, the more reliable the prediction to the inferred class. This is a way of establishing a trust factor for each prediction made by the VGG-16 model towards the inferred class. If the agreement between the filters is below a given threshold, the decision is considered to be a possible misclassification. 
The results of this analysis are presented in Table \ref{tab:1}. VGG-16 model was used to predict 5,764 test images of the CUB dataset \cite{WahCUB_200_2011}. Out of these, the model correctly predicted 69.5\% correctly, resulting in 1769 misclassifications. For each of the 5,764 test images, we use different metrics to compute the agreement between the activated filters and global MC filters for the inferred class. These include average recall, average F1 score, and thresholded recall. We use a skip threshold to skip high confidence prediction above the threshold. We use frequency threshold on the global MC filters to consider only those filters for comparison that have been activated at least the specified amount of times out of the maximum number of times a filter is activated for a particular class. Based on these conditions, we check whether a local prediction is likely to be misclassification if the specified metric score computed between the locally activated filters and global MC filters is below the average score threshold for the specified metric for the training set images. In Table \ref{tab:1}, using the average recall metric results in the successful identification of 31\% of the total misclassifications while 256 new misclassification are introduced due to the condition. And using the average F1 score metric, 38\% of the misclassifications are detected. 

\begin{table*}[]
\footnotesize
\centering
\caption{Automatic misclassification detection based on the similarity between local activated filters of each test image of CUB \cite{WahCUB_200_2011} dataset and global MC filters of respective inferred classes by VGG-16 model.}
\label{tab:1}
\begin{tabular}{|l|l|l|l|l|l|l|}
\hline
\textbf{Model} & \textbf{\begin{tabular}[c]{@{}l@{}}Total \\ errors\end{tabular}} & \textbf{\begin{tabular}[c]{@{}l@{}}Skip \\ threshold\end{tabular}} & \textbf{\begin{tabular}[c]{@{}l@{}}Freq. \\ threshold\end{tabular}} & \textbf{Metric} & \textbf{\begin{tabular}[c]{@{}l@{}}Errors \\ detected\end{tabular}} & \textbf{\begin{tabular}[c]{@{}l@{}}New \\ errors\end{tabular}} \\ \hline
\multirow{3}{*}{VGG-16} & \multirow{3}{*}{1769} & 90\% & 15\% & Avg. Recall & 541 (31\%) & 256 \\ \cline{3-7} 
 &  & 90\% & 15\% & Avg. F1 score & 680 (38\%) & 370 \\ \cline{3-7} 
 &  & 0\% & 0\% & Recall \textless 0.3 & 653 (37\%) & 369 \\ \hline
\end{tabular}
\end{table*}






\subsection{Model treatment and debugging}
\label{sec:four-4}
To treat the model weaknesses and reduce the number of misclassifications, we perform debugging to improve model performance. In this phase, we re-train the model with additional constraints to encourage MC filters activation for inferred classes and discourage non-MC filter activation for all other classes.
The model debugging results are presented in Table \ref{tab:2}. The table shows training and testing accuracies achieved when performing debugging using different $\lambda_1$ and $\lambda_2$ weights assigned to the loss maximizing the agreement with MC filters and minimizing agreement with non-MC filters as shown in Eq. \ref{eq:l_d}. The model debugging process is able to improve the test accuracy by almost 1.2\% from 69.48\% of the base model to 70.64\%.
To show the impact of the proposed optimisation, we compare the model improvement without incorporating the additional debugging constraints. In this default setting, re-training the base model in a similar way to debugging method, we are able to see an improved accuracy of 70.2\%, which is 0.4\% less than the debugging method. This shows that by encouraging MC filters activation, we are able to achieve a better model performance than fine-tuning the base model.






\begin{table}[]
\footnotesize
\centering
\caption{Model debugging performance using different loss weights and comparison with standard fine-tuning of the model.}
\label{tab:2}
\begin{tabular}{|c|l|l|l|l|}
\hline
\multicolumn{1}{|l|}{Model} & \textbf{\begin{tabular}[c]{@{}l@{}}MC filters \\ weight $\lambda_1$\end{tabular}} & \textbf{\begin{tabular}[c]{@{}l@{}}Non-MC \\ filters weight $\lambda_2$\end{tabular}} & \textbf{\begin{tabular}[c]{@{}l@{}}Train\\ acc. (\%)\end{tabular}} & \textbf{\begin{tabular}[c]{@{}l@{}}Test\\ acc. (\%)\end{tabular}} \\ \hline
\multicolumn{1}{|l|}{\textbf{VGG-16 base}} & - & - &  & 69.48 \\ \hline
\multicolumn{1}{|l|}{\textbf{Fine-tuned}} & - & - & 99.0 & 70.2 \\ \hline
\multirow{8}{*}{\textbf{\begin{tabular}[c]{@{}c@{}}VGG-16 \\ debugged\end{tabular}}} & 0.0001 & 0.00001 & 99.2 & 70.1 \\ \cline{2-5} 
 & 0.0002 & 0.00002 & 99.2 & 70.2 \\ \cline{2-5} 
 & 0.0005 & 0.00002 & 98.9 & 70.42 \\ \cline{2-5} 
 & 0.0005 & 0.00005 & 99.1 & 70.2 \\ \cline{2-5} 
 & 0.001 & 0.00002 & 98.5 & 70.61 \\ \cline{2-5} 
 & 0.002 & 0.00002 & 97.5 & 70.4 \\ \cline{2-5} 
 & 0.001 & 0.00005 & 98.7 & 70.64 \\ \cline{2-5} 
 & 0.001 & 0.0001 & 99.1 & 70.57 \\ \hline
\end{tabular}
\end{table}


\begin{table}[]
\footnotesize
\centering
\caption{Changes in class recall resulting from model debugging method.}
\label{tab:my-table-improvements}
\begin{tabular}{|llll|}
\hline
\multicolumn{1}{|l|}{\textbf{Class}} & \multicolumn{1}{l|}{\textbf{Original recall}} & \multicolumn{1}{l|}{\textbf{Debugged recall}} & \textbf{Change} \\ \hline
\multicolumn{4}{|c|}{Top 3 classes with improved recall}                                                       \\ \hline
\multicolumn{1}{|l|}{Boat\_tailed\_Grackle}    & \multicolumn{1}{l|}{0.30} & \multicolumn{1}{l|}{0.70} & 0.40  \\ \hline
\multicolumn{1}{|l|}{Marsh\_Wren}              & \multicolumn{1}{l|}{0.57} & \multicolumn{1}{l|}{0.77} & 0.20  \\ \hline
\multicolumn{1}{|l|}{Pomarine\_Jaeger}         & \multicolumn{1}{l|}{0.37} & \multicolumn{1}{l|}{0.57} & 0.20  \\ \hline
\multicolumn{4}{|c|}{Top 3 classes with decreased recall}                                                      \\ \hline
\multicolumn{1}{|l|}{American\_Crow}           & \multicolumn{1}{l|}{0.50} & \multicolumn{1}{l|}{0.30} & -0.20 \\ \hline
\multicolumn{1}{|l|}{Black\_footed\_Albatross} & \multicolumn{1}{l|}{0.67} & \multicolumn{1}{l|}{0.50} & -0.17 \\ \hline
\multicolumn{1}{|l|}{Chuck\_will\_Widow}       & \multicolumn{1}{l|}{0.69} & \multicolumn{1}{l|}{0.54} & -0.15 \\ \hline
\end{tabular}
\end{table}

\subsection{Qualitative analysis}
\label{sec:four-5}
This section provides a qualitative analysis of the model debugging method based on finding classes for which the accuracy improved or decreased and visualizing cases from these classes. Table \ref{tab:my-table-improvements} shows the top 3 classes with the most improved and decreased class recall scores resulting from the model debugging process. It is seen that a significant improvement of 40\% recall is observed for the `Boat-tailed Grackle' class, while the `American Crow' class is the worst affected with a decrease of 20\%. The reason for the decrease in class recall could be that initially, the model relies on biases to achieve the specified accuracy that may be mistakenly classified to the correct class. Now, when we perform model debugging and push the model to use class-specific filters to improve model accuracy, the biases that the model was using have been reduced, which led to a decrease in the class recall of some of the classes on test data.

\begin{figure*}[bt!]
    \centering
    
    \subfloat[Correct with 63.5\% confidence] {\includegraphics[width=.25\textwidth]{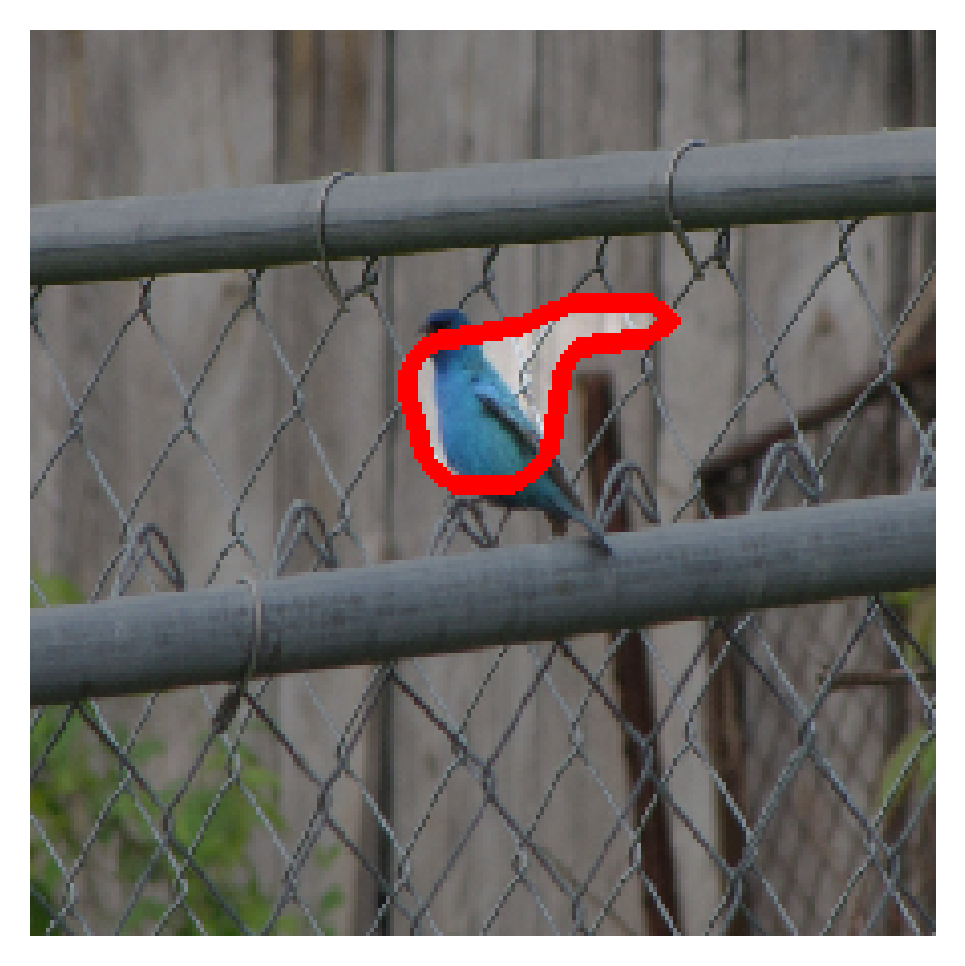} }
    \subfloat[Misclassification] {\includegraphics[width=.25\textwidth]{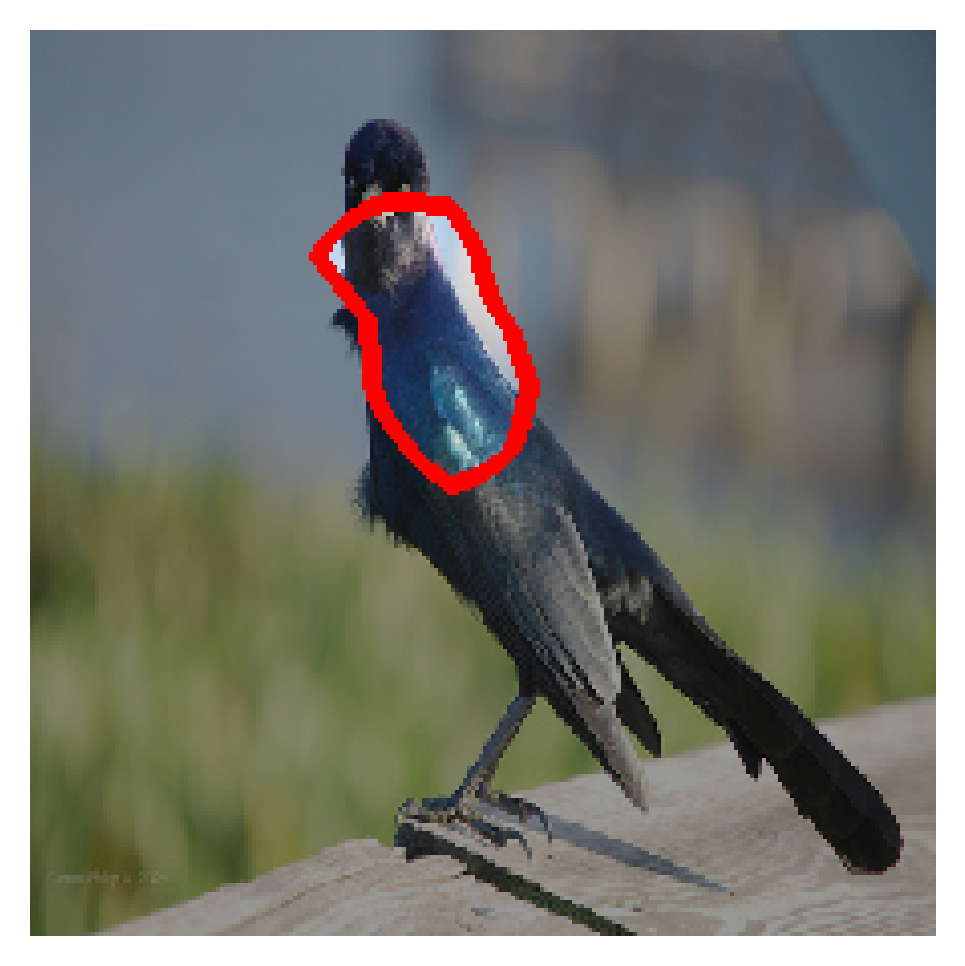} }
    \subfloat[New error] {\includegraphics[width=.25\textwidth]{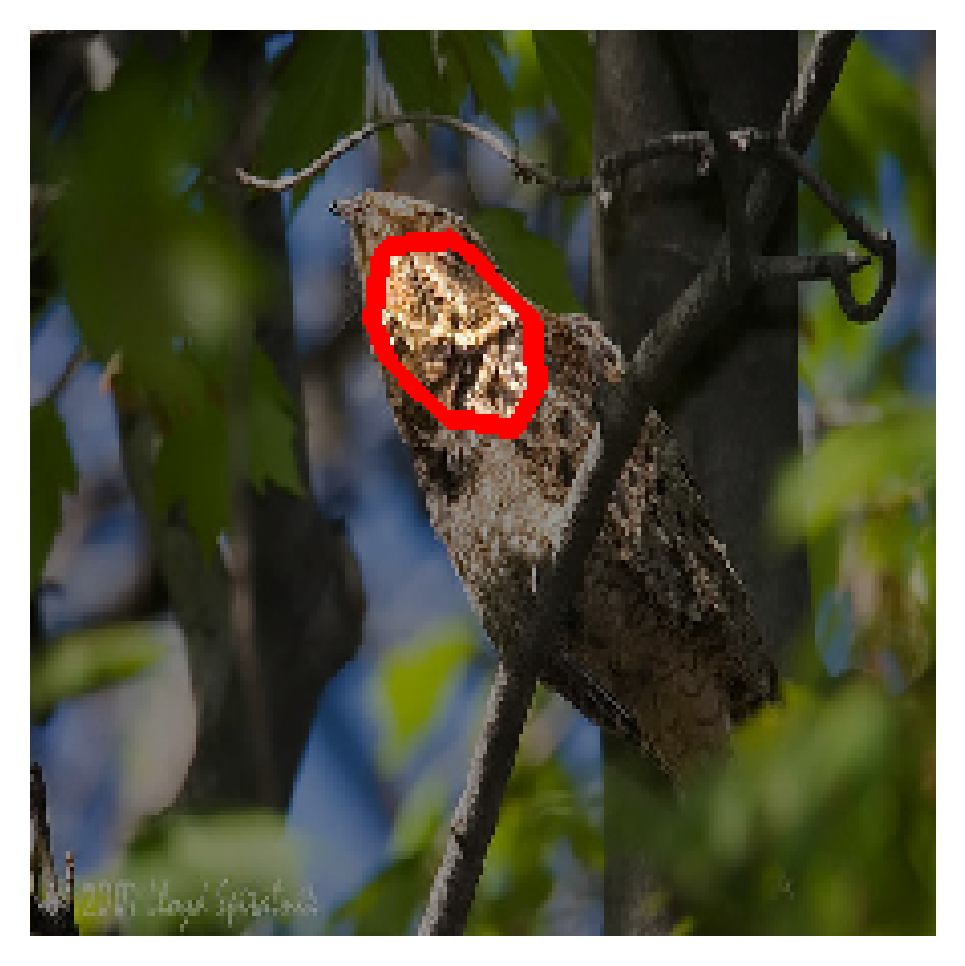} }
    
    \subfloat[Confidence improved to 95.0\%] {\includegraphics[width=.25\textwidth]{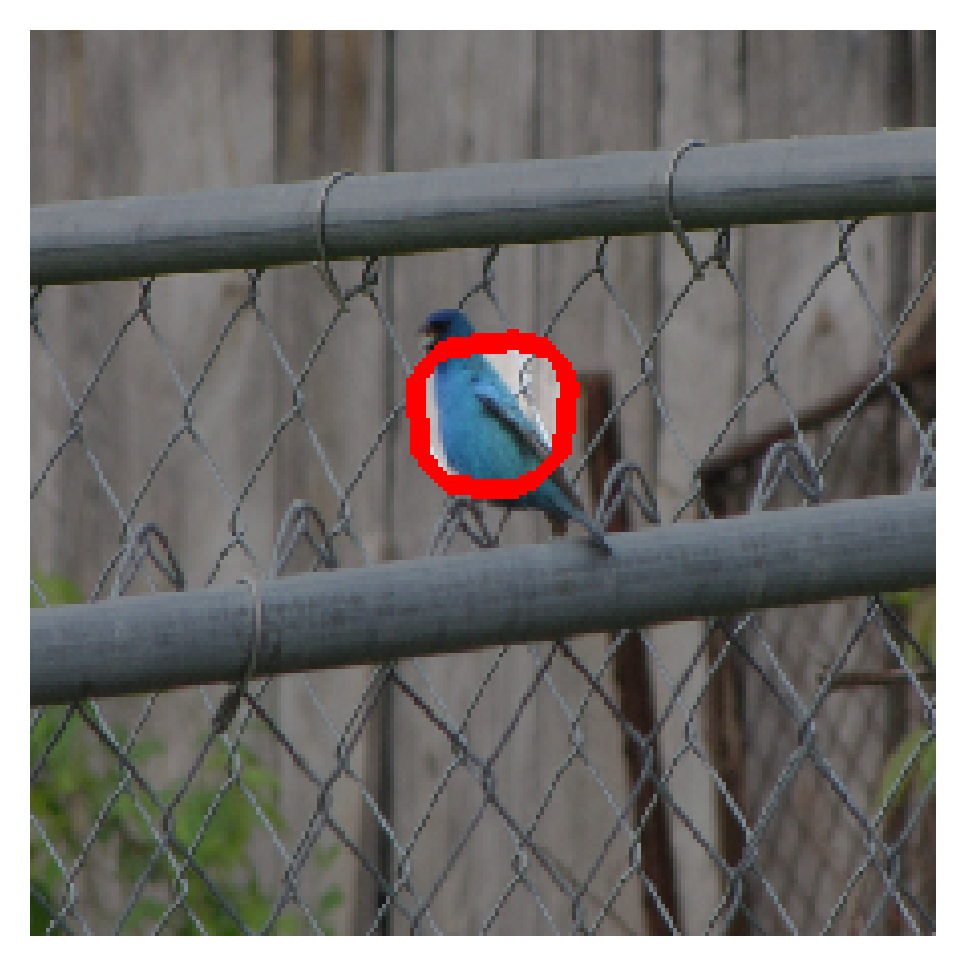} }
    \subfloat[Error corrected ] {\includegraphics[width=.25\textwidth]{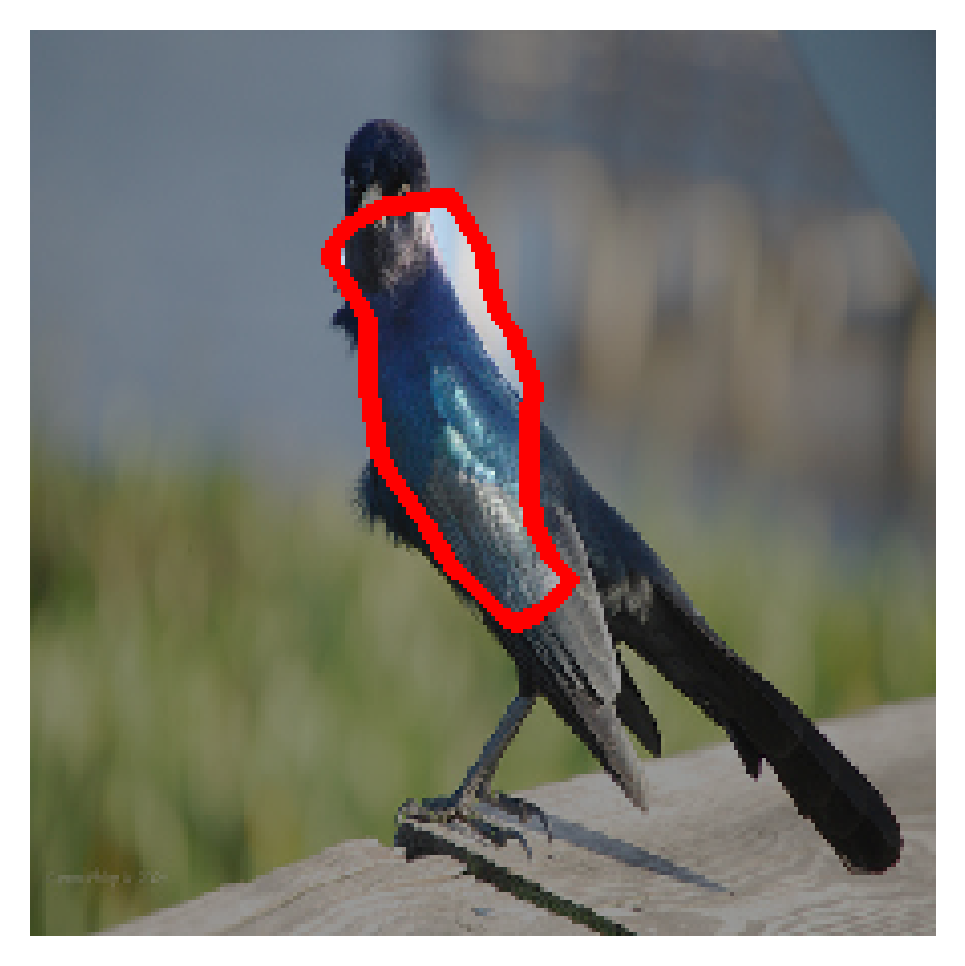} }
    \subfloat[Sample from incorrect class] {\includegraphics[width=.25\textwidth]{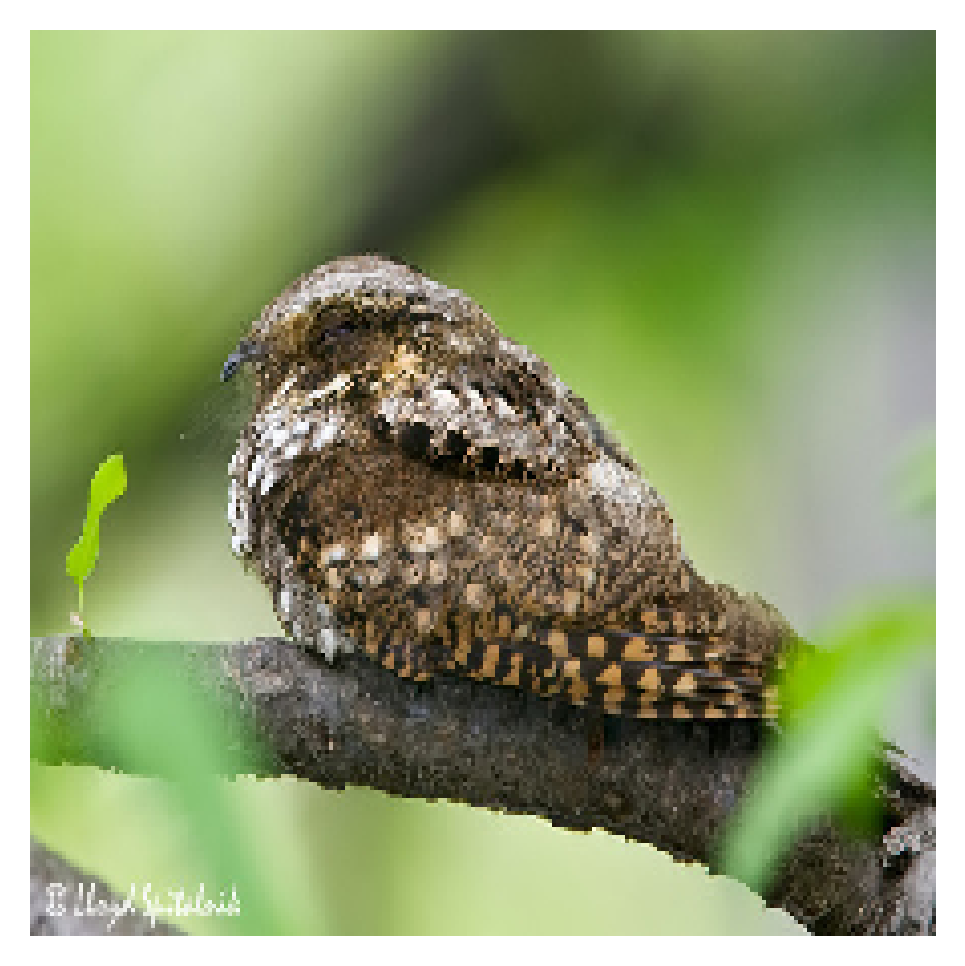} }
        
    \caption{Success and failure cases of the proposed method. In (a), the pre-trained VGG-16 model classified an image correctly but with low confidence of 63.5\%. The improved VGG-16 improved the confidence to 95\% in (d). In (b), the original model made a misclassification of the `Boat-tailed Grackle' class image to the `Brewer Blackbird' class, which is fixed by the improved model in (e). In (c), the improved model made a wrong prediction to the `Whip Poor Will' class with highlighted features detected that are similar to the sample of the incorrect class shown in (f).} 
        \label{fig:visualizations}
\end{figure*}

Figure \ref{fig:visualizations} shows some visual results of the proposed method. Figure parts (a) and (d) show a case where the original VGG-16 model was less confident on a prediction, and the proposed method resulted in improved confidence as indicated by the highlighted portion (computed using GradCAM \cite{selvaraju2017grad} on the image where the focus of the model is improved on the bird and shifted from the background region. In parts (b) and (e), the proposed model resulted in a mistake being corrected from the original VGG-16 model. While parts (c) and (f) show a failure case of the proposed method where the original model was correct, the debugged model resulted in a new error. The reason for this mistake is that the region highlighted in the mistaken classification in (c) shows a texture that is similar to the distinguishing features of the sample image from the wrong class of `Whip Poor Will'.





\subsection{Discussion and limitations}
\label{sec:four-6}

Based on the misclassification analysis presented in section \ref{sec:four-2}, we can conclude that although a lot of possible misclassifications are identified, many are still not identified. This suggests that filters activated in high-confidence test cases are not well aligned with global MC filters identified from the training set. This, in turn, indicates that the model is overfitting on training data, which is also evident from the large gap in training and testing accuracy of the VGG-16 model on the CUB dataset. Furthermore, results in section \ref{sec:four-4} demonstrated that by encouraging the model to predict the class-relevant filters and ignoring other filters, it is possible to improve the accuracy of the model by aligning the decision-making process towards class-specific concepts identified using counterfactual reasoning.

One of the drawbacks of the proposed method is that its ability to identify class-relevant important filters used for aligning model decisions and debugging is dependent on how accurate the original model was on the classification task. If the model was less accurate, then the MC filters would be less faithful to the model's decision-making process. Moreover, if there is a large difference between the model's training and testing accuracy, then the MC filters extracted from the training set will be less aligned and less useful for judging misclassification from the test set.








%% file: conclusion.tex
\section{Conclusion}
\label{sec:five}

This paper presented a method to improve the performance of CNN-based image classifiers and perform misclassification analysis utilizing explainability methods. We proposed a novel approach that leverages counterfactual concepts by identifying crucial filters used in the decision-making process and retraining models to encourage the activation of class-relevant important filters and discourage the activation of irrelevant filters. Through this process, we minimized the deviation of activation patterns, aligning local predictions with the global activation patterns of their respective inferred classes. Experimental results on publicly available datasets confirmed the effectiveness of our approach, with an observed improvement of 1-2\%.
